\documentclass[letterpaper,10pt,conference]{ieeeconf}
\IEEEoverridecommandlockouts       
\usepackage{amssymb}
\usepackage{amsmath}
\usepackage[english]{babel}
\usepackage{float}
\usepackage{graphicx}
\usepackage[hidelinks]{hyperref}
\usepackage{mathtools}

\usepackage{filecontents}
\usepackage[noadjust]{cite}
\usepackage[font=footnotesize,labelfont=footnotesize]{subcaption}
\usepackage[font=footnotesize,labelfont=footnotesize]{caption}
\usepackage{comment}
\usepackage{authblk}
\usepackage{multirow}
\usepackage{xcolor}
\usepackage{algorithm}
\usepackage{algorithmic}
\usepackage{xspace}
\usepackage{hhline}

\usepackage{tabularx}
\usepackage{makecell}
\usepackage{subcaption}

\definecolor{darkblue}{rgb}{0.15,0.15,0.55}
\definecolor{lightgrey}{rgb}{0.75,0.75,0.75}

\newcommand{\thm}{\noindent \textbf{Theorem}\xspace}

\newcommand{\pf}{\noindent \textbf{Proof}\xspace}

\newcommand{\qed}{\hfill $\square$}

\begin{document}
\title{\LARGE \bf Stop-N-Go: Search-based Conflict Resolution for Motion Planning of Multiple Robotic Manipulators\vspace{-7pt}}
\author{Gidon Han$^{1}$, Jeongwoo Park$^{1}$, and Changjoo Nam$^{1,*}$%
\thanks{This work was supported by the National Research Foundation of Korea (NRF) grants funded by the Korea government (MSIT) (No. 2022R1C1C1008476 and No. RS-2024-00411007). $^1$Dept. of Electronic Engineering at Sogang University, Seoul, Korea.  $^*$\textit{Corresponding author: Changjoo Nam} ({\tt\small cjnam@sogang.ac.kr})}
}

\maketitle

\begin{abstract}
We address the motion planning problem for multiple robotic manipulators in packed environments where shared workspace can result in goal positions occupied or blocked by other robots unless those other robots move away to make the goal positions free. While planning in a coupled configuration space (C-space) is straightforward, it struggles to scale with the number of robots and often fails to find solutions. Decoupled planning is faster but frequently leads to conflicts between trajectories.

We propose a conflict resolution approach that inserts pauses into individually planned trajectories using an A$^*$ search strategy to minimize the makespan--the total time until all robots complete their tasks. This method allows some robots to stop, enabling others to move without collisions, and maintains short distances in the C-space. It also effectively handles cases where goal positions are initially blocked by other robots. Experimental results show that our method successfully solves challenging instances where baseline methods fail to find feasible solutions.
\end{abstract}\vspace{-3pt}

\section{Introduction}
\label{sec:intro}


Multi-manipulator systems, such as those in manufacturing, require efficient motion planning for tasks that demand close coordination, like the simultaneous assembly of parts. High degree-of-freedom (DOF) manipulators must find collision-free trajectories, but planning in the composite configuration space (C-space) of all robots (i.e., coupled motion planning) is computationally demanding. While sampling-based planners like RRT-connect~\cite{RRTconnect} can approximate the search space efficiently, advanced algorithms like dRRT$^*$~\cite{dRRT} tackle this challenge with asymptotic optimality.

\begin{figure}[t]
    \centering
    \captionsetup{skip=-8pt}
    \begin{subfigure}{0.312\textwidth}
        \includegraphics[width=\textwidth]{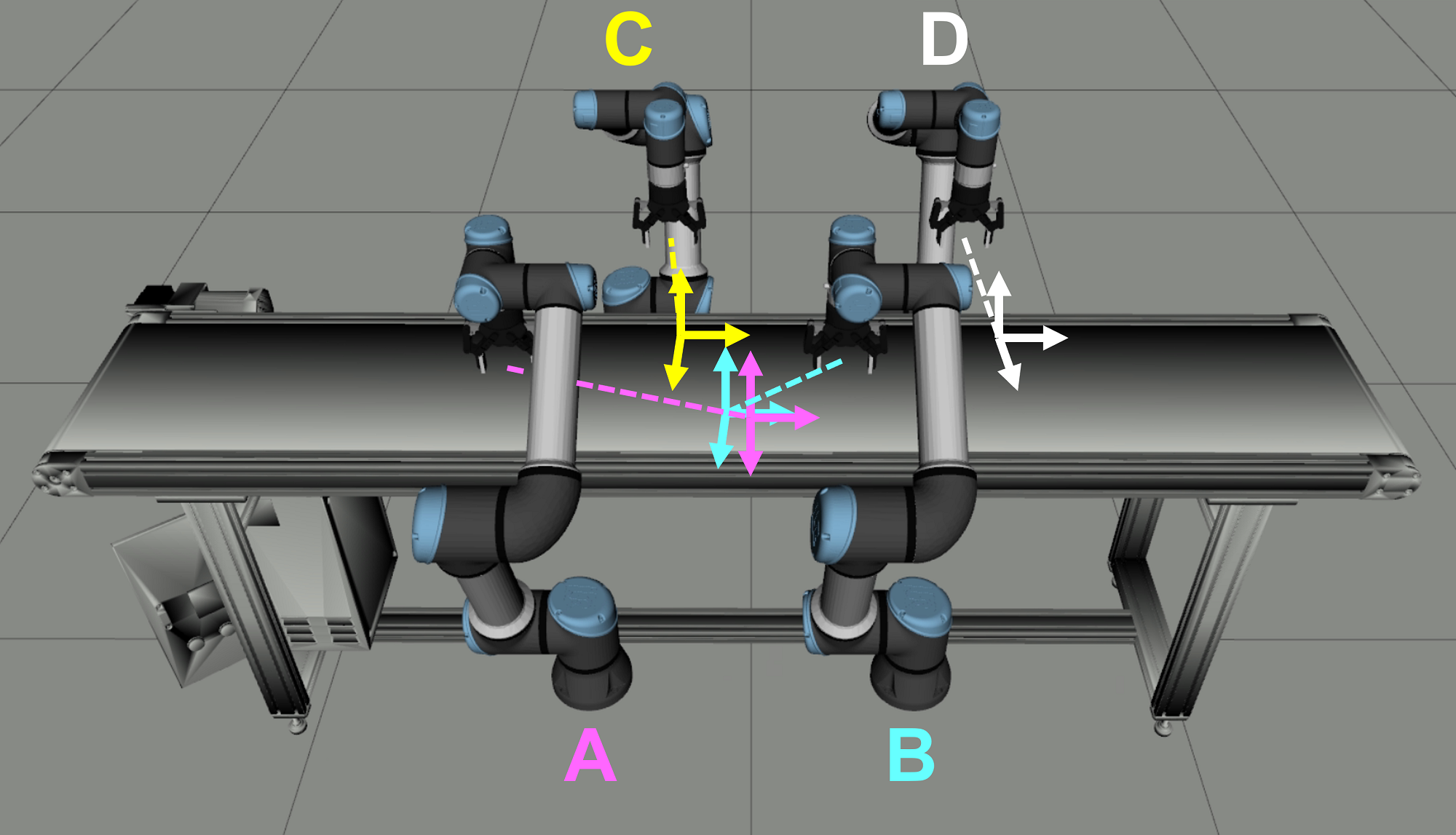}
        \label{fig:bolt}
    \end{subfigure}\,%
    \begin{subfigure}{0.185\textwidth}
        \includegraphics[width=\textwidth]{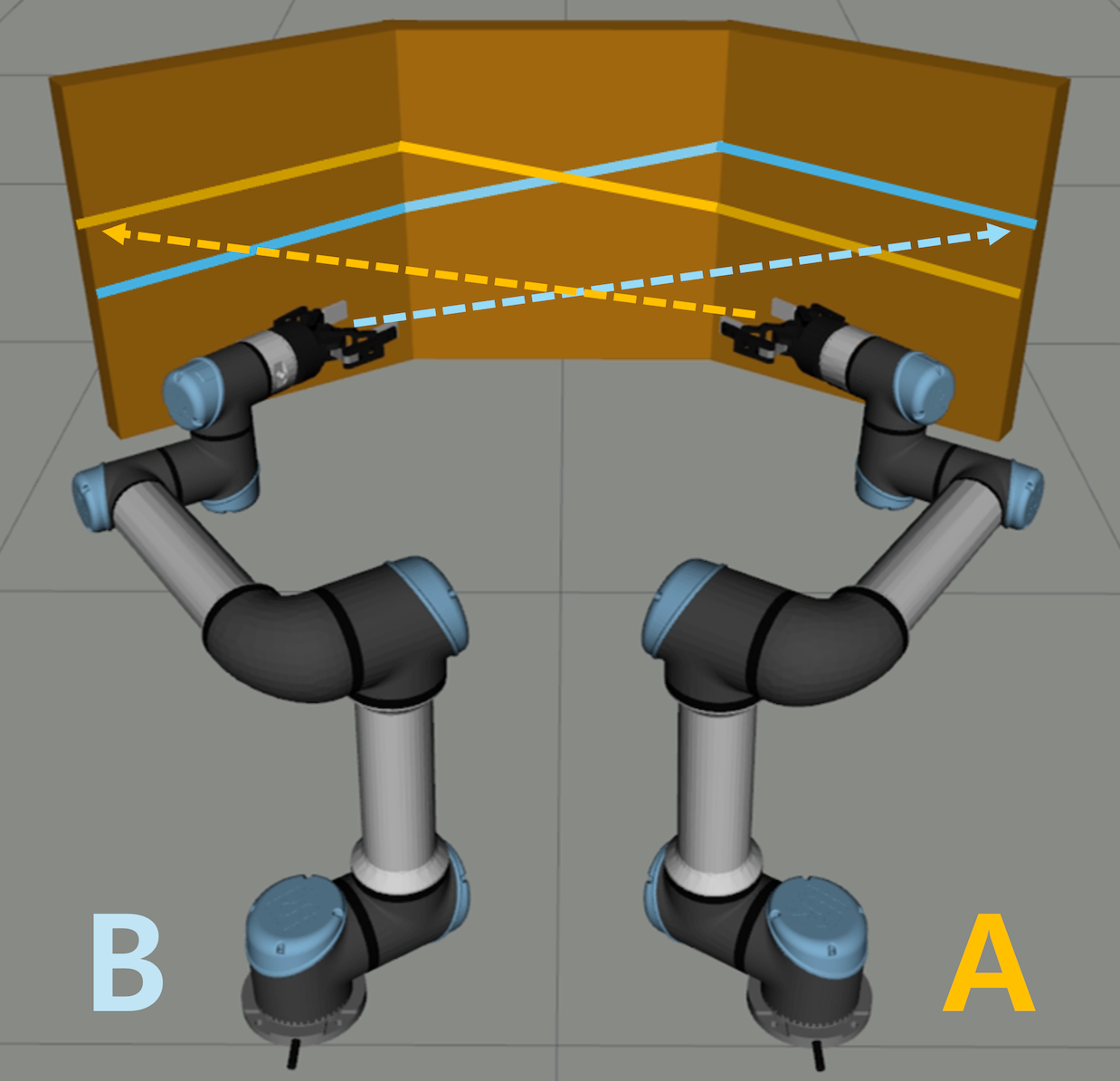}
        \label{fig:weld}
    \end{subfigure}
\caption{Example tasks where robots are assigned specific task locations and need to return after completing their tasks. (Left) Robots \textsf{A} and \textsf{B} have overlapping goal positions to pick. (Right) Robots need to follow predefined trajectories (for welding or painting tasks) but their trajectories block each other to move.}
\label{fig:ex}\vspace{-20pt}
\end{figure}

However, some planning problems require consideration of the temporal dimension. For example, robots performing a bolting task on a conveyor must temporarily reach and then leave their goal positions (left of Fig.~\ref{fig:ex}). Robots performing welding and spraying tasks should follow predefined end-effector paths (right of Fig.~\ref{fig:ex}). In dense environments with narrow free C-space, some robots may be unable to achieve feasible configurations unless other robots vacate their goal positions as illustrated in Fig.\ref{fig:ex}.


Motion planning methods that neglect the temporal dimension struggle to find conflict-free trajectories. For instance, while dRRT$^*$ can optimize trajectories in high-dimensional spaces, it does not account for the timing of robot movements, failing in scenarios requiring temporal coordination. Similarly, Prioritized Planning (PP)~\cite{PP} reduces the search space by assigning priorities but ignores timing, leading to infeasible trajectories when higher-priority robots block lower-priority goals. Executing tasks in turn could be a solution method if robots wait for minimal duration for others to move away. However, it is a P-SPACE hard problem to plan for the optimal timing and trajectories to minimize the total time elapsed until all robots finish their tasks (i.e., \textit{makespan})~\cite{stilman2005navigation}.


We propose a method called \textit{Stop-N-Go}, which determines when to stop and resume execution of trajectories to minimize the makespan. Our method inserts pauses into individually planned trajectories to avoid collisions without altering the spatial dimensions of the trajectories. Without explicitly planning in the C-space augmented by the temporal dimension, our method resolves trajectory conflicts by adjusting their timestamps. Specifically, our method performs an A$^*$ search to find collision-free trajectories with the minimum makespan by inserting pauses into the trajectories to avoid collisions.

Our contributions include: (i) developing a method to find conflict-free trajectories, (ii) ensuring probabilistic completeness and optimality, and (iii) conducting extensive experiments, including comparisons with baseline methods.

\section{Related Work}
\label{sec:related}

In multi-manipulator systems, motion planning often relies on coupled approaches. Probabilistic Roadmaps (PRMs)~\cite{PRM} are popular for high-dimensional C-spaces. However, they become computationally expensive in cluttered environments. dRRT$^*$\cite{dRRT} treats the C-spaces of all robots as a single composite space, but it struggles in packed environments. In instances like those in Fig.\ref{fig:ex}, robots \textsf{A} and \textsf{B} fail to find solutions in the composite C-space due to limited free space. They could reach their goals by adjusting arrival times, but adding a temporal dimension results in synchronized waits for all robots, negating individual adjustments.

Two-level approaches have been widely developed to resolve conflicts by separating high-level conflict detection from low-level trajectory adjustments. Conflict-Based Search (CBS)~\cite{CBS} identifies conflicts at the high level and applies constraints at the low level. Prioritized Planning (PP)~\cite{PP} assigns priorities, requiring lower-priority robots to avoid higher-priority ones. While these methods can adjust for spatial and temporal conflicts by augmenting the C-space with time, no current CBS or PP variants effectively handle continuous C-space and time in multi-manipulator setting.

Reinforcement learning (RL)-based motion generation methods have recently emerged. In~\cite{ha2020multiarm}, a Soft Actor-Critic (SAC)~\cite{soft_actor_critic} agent is pretrained using expert demonstrations from BiRRT\cite{BiRRT} and fine-tuned for various scenarios. Unlike traditional methods, RL can generate motions quickly even with multiple robots. However, it also struggles with scenarios where goal positions are temporarily overlapped, as demonstrated in our experiments.

As described, existing methods have shown limitations in solving the planning instances with temporal overlaps in the goal positions. We observe a need of handling temporality of trajectories in the continuous C-space for multi-manipulator motion planning. We aim to bring an advancement that enables temporal adjustments to resolve conflicts without explicit spatio-temporal motion planning, which is complex to implement and computationally expensive. 

\section{Problem Description}  
\label{sec:prob}


We consider $k$ robots $R_i$ for $i = 1, \cdots, k$. Suppose that $R_i$ is assigned to perform a sequence of tasks whose goal positions are given by $\mathcal{G}_i=(g_i^{1}, g_i^{2}, \ldots, g_{i}^{n_i})$ where $n_i$ is the number of tasks assigned to $R_i$. We assume that $g_i^{1} = g_{i}^{n_i}$ which means that the robot comes back to its initial position once it finishes all assigned tasks. Each goal position can be occupied by other robots temporarily but should have a moment that it becomes free so the robot can reach the goal to perform the task.

The goal is to find trajectories of all robots $\mathcal{T} = \{\mathcal{T}_1, \cdots, \mathcal{T}_k\}$ to finish their task sequences without collisions while minimizing the time to finish all tasks of all robots. Each trajectory of $R_i$ is a sequence of tuples $\mathcal{T}_i = ((q_i^1, 1), \cdots (q_i^j, t), \cdots)$ where $q_i^j$ is the $j$-th configuration of $R_i$ at $t$. At $t = \tau_i$, $R_i$ returns to its initial position after finishing all tasks in $\mathcal{G}_i$. Therefore, we want to find $\mathcal{T}$ with the minimum $\max \tau_i$ for $i=1, \cdots, k$, which is the \textit{makespan}. This problem is equivalent to Navigation Among Movable Obstacles (NAMO) which is P-SPACE hard~\cite{stilman2005navigation}.

\section{Stop-N-Go: Temporal Adjustment for Multi-robot Motion Planning}

\subsection{Overview}
The input to our method is a set of initial trajectories $\mathcal{T}_{\text{init}}$ for all robots. The trajectories can be generated individually using any standard motion planner (e.g., \cite{RRTconnect}) by treating other robots as static obstacles. Our method modifies these trajectories by inserting pauses between configurations. For example, $\mathcal{T}_1 = ((q_1^1, 1), (q_1^2, 2), (q_1^3, 3))$ may become $((q_1^1, 1), (q_1^2, 2), (q_1^2, 3), (q_1^3, 4))$, causing the robot to stop between $t=2$ and $t=4$. Notice that time steps are not necessarily integers.

\begin{figure}[t]
    \vspace{-5pt}
    \captionsetup{skip=0pt}
    \centering
    \includegraphics[width=0.48\textwidth]{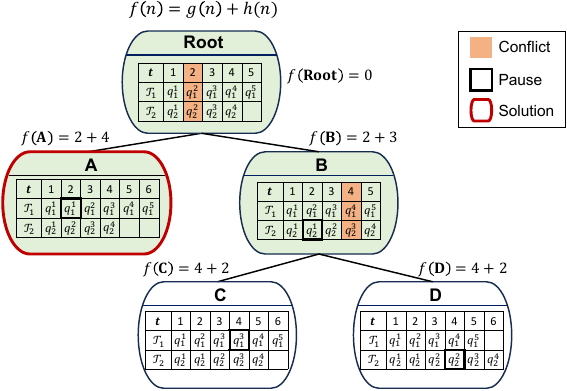}
    \caption{An overview of the proposed method. For each conflict found, two successor nodes are generated to resolve the conflict by inserting pauses. Node \textsf{A} is found to be conflict-free so the search terminates. The nodes shaded with green represent expanded nodes where the one with red bold outline is the solution.}
    \label{fig:overview}\vspace{-10pt}
\end{figure}

The method uses A$^*$ to search to find the modified trajectories $\mathcal{T}$ with the minimum makespan. The cost function combines the time until the current conflict (i.e., path cost) and the remaining time to complete all tasks (i.e., heuristic cost), equal to the estimated makespan. Goal testing occurs when a node is expanded, with ties broken lexicographically. Fig.~\ref{fig:overview} shows an example where $\mathcal{T}_{\text{init}}$ for robots $i=1,2$ conflicts at $t=2$. The root node is expanded, generating two successor nodes where each node represents the trajectories where one of the robots modifies its trajectory by inserting a pause. In the worst case, $R_i$ should wait all the other robots finish their trajectories. However, there are chances to reduce the waiting time where our method provides an effective way for it, described in the following sections.

\subsection{The Stop-N-Go algorithm}
\label{sec:method1}
We describe the proposed method with details with the pseudocode shown in Alg.~\ref{alg:stopngo}. It receives the individually found trajectories of all $k$ robots $\mathcal{T}_{\text{init}}$ and produces the collision-free trajectories with temporal adjustments $\mathcal{T}$. 

\begin{figure}[t]
    \centering
    \captionsetup{skip=0pt}
    \includegraphics[width=1\columnwidth]{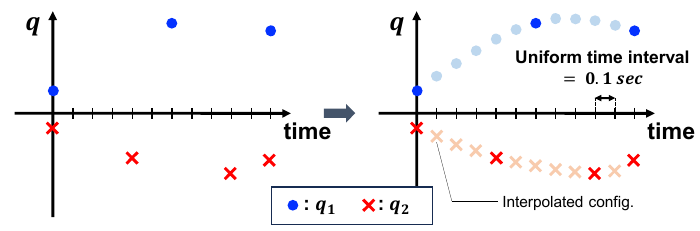}
\caption{Interpolation of the initial trajectories for synchronization. Since the C-space of high-DOF robots cannot be visualized, we use the one-dimensional C-space ($q_i$ for $R_i$). (Left) Tthe initial trajectories with non-uniform time intervals. (Right) After interpolation with $t_\text{intv} = 0.1$.}
\label{fig:interpolate}
 \vspace{-5pt}
\end{figure}

The $k$ trajectories in $\mathcal{T}_{\text{init}}$ have unsynchronized time steps since they are planned individually, and configurations are not uniformly distributed over time. For example, $\mathcal{T}_1 = ((q_1^1, 0), (q_1^2, 0.6), (q_1^3, 1.1))$ and $\mathcal{T}_2 = ((q_2^1, 0), (q_2^2, 0.4), (q_2^3, 0.9), (q_2^4, 1.1))$. To address this, we synchronize all trajectories to a common time interval, $t_{\text{intv}}$, via interpolation in line~\ref{alg1:interpolate} of Alg.\ref{alg:stopngo}, as exemplified in Fig.\ref{fig:interpolate}.


Our A$^*$ search starts with the synchronized trajectories (line~\ref{alg1:root}) to minimize the makespan. The evaluation function for a node $\mathcal{N}$ is:
\begin{equation}
    f(\mathcal{N}) = g(\mathcal{N}) + h(\mathcal{N})
\end{equation}
where $g(\mathcal{N})$ is the path cost (i.e., execution time until $\mathcal{N}$), and $h(\mathcal{N})$ is the heuristic (i.e., remaining time to complete all tasks). The makespan estimate is the longest trajectory length, in other words, $\max \tau_i$ for all $R_i$. Line~\ref{alg1:get} selects the node with the lowest $f$ in the frontier $OPEN$. The trajectories are scanned iteratively (lines~\ref{alg1:conflict_s}--\ref{alg1:conflict_e}) to detect conflicts\footnote{Any collision checker can be used while we use FCL~\cite{FCL} with Assimp (Open Asset Import Library) for importing robot models.}, and conflicts are resolved in \textsc{ConflictResolution}. If no conflict is found, the algorithm terminates (line~\ref{alg1:terminate}) and sends the final trajectories to the controller.


If a conflict is detected, two successor nodes are generated for $R_i$ and $R_j$, where one of the trajectories is modified by inserting pauses to avoid the conflict (lines~\ref{alg1:successor_s}--\ref{alg1:successor_e}). The search tree branches based on pairwise conflicts, and conflicts are resolved by modifying $\mathcal{T}_i$ (line~\ref{alg1:conflict_i}) and $\mathcal{T}_j$ (line~\ref{alg1:conflict_j}). Since the resulting $\mathcal{T}$ still can have conflicts, the successor nodes with the temporal modification are inserted to $OPEN$ until termination.


\begin{algorithm}[t]
\small
\caption{\textsc{Stop-N-Go}}
\label{alg:stopngo}
\begin{algorithmic}[1]
\renewcommand{\algorithmicrequire}{\textbf{Input:} }
\renewcommand{\algorithmicensure}{\textbf{Output: }}
\REQUIRE $\mathcal{T}_{\text{init}} = \{\mathcal{T}_1, \dots, \mathcal{T}_k\}$, $jump$
\ENSURE $\mathcal{T}$

\STATE $OPEN \gets \emptyset$, $\mathcal{N}.\mathcal{T} \gets \emptyset$, $\mathcal{N}.cost \gets 0$, $\mathcal{N}.t_s \gets 0$, $\mathcal{C}\gets \emptyset$
\STATE $\mathcal{N}.\mathcal{T} \gets$ \textsc{SetClock}$(\mathcal{T}_{\text{init}})$ \label{alg1:interpolate} 
\STATE \text{Insert} $\mathcal{N}$ \text{into}\label{alg1:root} $OPEN$

\WHILE{\text{$OPEN$ is not empty}}
    \STATE $\mathcal{N} \gets \textsc{GetBestNode}(OPEN)$\label{alg1:get}
    \FOR{$t \gets \mathcal{N}.t_s$ \textbf{to} $\max(\tau)$}\label{alg1:conflict_s} 
        \STATE $\mathcal{C}, \mathcal{N}.t_s \gets \textsc{ConflictCheck}(\mathcal{T}, t)$ \label{alg1:conflict}
    \ENDFOR \label{alg1:conflict_e}
    \IF{$\mathcal{C} = \emptyset$} \label{alg1:done_s}
        \RETURN $\mathcal{N}.\mathcal{T}$ \label{alg1:terminate}
    \ELSE \label{alg1:successor_s} 
        \STATE $(i, j) \gets \mathcal{C}$
        \STATE $\mathcal{N}^{\prime} \gets$ \textsc{copy}($\mathcal{N}$)
        \STATE $\mathcal{N}^{\prime} \gets$ \textsc{ConflictResolution}($\mathcal{N}^{\prime}$, $\mathcal{C}$, $R_i$, $jump$)\label{alg1:conflict_i}
        \STATE $\mathcal{N}^{\prime}.cost \gets$ \textsc{CalculateCost}$(\mathcal{N}^{\prime}.\mathcal{T})$
        \STATE \text{Insert} $\mathcal{N}^{\prime}$ \text{into $OPEN$}
        \STATE $\mathcal{N}^{\prime} \gets$ \textsc{copy}($\mathcal{N}$)
        \STATE $\mathcal{N}^{\prime} \gets$ \textsc{ConflictResolution}($\mathcal{N}^{\prime}$, $\mathcal{C}$, $R_j$, $jump$) \label{alg1:conflict_j}
        \STATE $\mathcal{N}^{\prime}.cost \gets$ \textsc{CalculateCost}$(\mathcal{N}^{\prime}.\mathcal{T})$
        \STATE \text{Insert} $\mathcal{N}^{\prime}$ \text{into $OPEN$}
    \ENDIF \label{alg1:successor_e} 
\ENDWHILE
\RETURN \text{Failure}
\end{algorithmic}
\end{algorithm}

\subsection{Conflict Resolution}
\label{sec:conflict}

Conflicts are resolved by inserting pauses into the trajectory of a robot, allowing one robot to wait while the other moves without adjustments. Alg.~\ref{alg:resolution} shows the procedure implemented in \textsc{ConflictResolution}. Temporal adjustments involve finding when to stop and resume motion by (i) identifying the stop time $t_p$, (ii) obtaining the configuration $q$ of the robot at $t_p$, and (iii) determining the pause duration.


The variable $n$ indicates how far back in the trajectory the robot should pause to resolve a conflict, starting at $1$ (line~\ref{alg2:n}) and incrementing each time a conflict occurs after the $n$-th previous motion (line~\ref{alg2:n_incre}). Line~\ref{alg2:search} finds $t_p$ in $\mathcal{T}.R$ (i.e., the trajectory of $R$) by searching for the $n$-th largest $t$ where $t < t_s$ and $q \neq q^\star$, with $(q, t_p)$ and $(q^\star, t_s)$ being the respective configurations and time steps. The trajectory is then shifted backward by $t_s - t_p$, and $q$ at $t_p$ is copied to fill the resulting empty slots (line~\ref{alg2:insert1}).

\begin{figure}[t]
    \centering
    \captionsetup{skip=0pt}
    \includegraphics[width=0.4\textwidth]{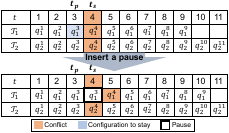}
    \caption{An illustration of how a detected conflict is resolved by inserting a pause at $t_s = 4$. The conflict is resolved by shifting the trajectory and adding a pause, which allows $\mathcal{T}_2$ to proceed without conflict.}
    \label{fig:conflict_resolution}\vspace{-5pt}
\end{figure}

In Fig.~\ref{fig:conflict_resolution}, a conflict at $t_s = 4$ is highlighted in red. If $R = R_1$, \textsc{SearchPauseStep} finds a previous configuration $q$ different from $q^\star = q_1^4$. If $q = q_1^3$ (blue), shifting the trajectory by $t_s - t_p = 1$ and inserting $q$ at $t = 4$ makes $R$ wait in the same configuration from $t = 3$ to $4$. Note that $t_p$ is not always $t_s - 1$ as some configurations may repeat.


Even after inserting pauses in $\mathcal{T}.R$, additional conflicts may arise due to the shift. We resolve these conflicts up to $t_s$ by scanning the trajectories between $t_p$ and $t_s$. If another conflict is found (i.e., $\mathcal{C} \neq \emptyset$), $n$ is incremented, and the process repeats until no conflicts remain.


\begin{algorithm}[t]
\small
\caption{\textsc{ConflictResolution}}
\label{alg:resolution}
\begin{algorithmic}[1]
\renewcommand{\algorithmicrequire}{\textbf{Input:} }
\renewcommand{\algorithmicensure}{\textbf{Output: }}
\REQUIRE $\mathcal{N}$, $\mathcal{C}$, $R$, $jump$
\ENSURE $\mathcal{N}$
\STATE $n \gets 1$  \label{alg2:n}
\WHILE{$\mathcal{C} \neq \emptyset$} \label{alg2:while_s}
    \STATE $\mathcal{T} \gets \mathcal{N}.\mathcal{T}$, $t_s \gets \mathcal{N}.t_s$

    \STATE $(q, t_p) \gets \textsc{SearchPauseStep}(\mathcal{T}.R, t_s, n)$ \label{alg2:search}
    \STATE $\textsc{Insert}(\mathcal{T}.R, q, t_s, t_s-t_p)$ \label{alg2:insert1}
    \FOR{$i \gets t_p$ \textbf{to} $t_s$}\label{alg2:conflict_s} 
        \STATE $\mathcal{C}, t_s \gets \textsc{ConflictCheck}(\{\mathcal{T}.\mathcal{C}[0] ,\mathcal{T}.\mathcal{C}[1] \}, i)$
    \ENDFOR \label{alg2:conflict_e}
    \STATE $n \gets n + 1$  \label{alg2:n_incre}
\ENDWHILE\label{alg2:while_e}
\IF{$jump$}\label{alg2:jump}
    \STATE $t_c \gets \textsc{SearchJumpStep}(\{\mathcal{T}.\mathcal{C}[0],\mathcal{T}.\mathcal{C}[1] \}, R, t_s)$   
    \STATE $(q, t_s) \gets \mathcal{T}.R[t_s]$
    \STATE $\textsc{Insert}(\mathcal{T}.R, q, t_c, t_c-t_s)$
\ENDIF
\STATE $\mathcal{N}.\mathcal{T} \gets \mathcal{T}$\label{alg2:return1}
\RETURN $\mathcal{N}$\label{alg2:return2}
\end{algorithmic}
\end{algorithm}

Conflicts may also occur after $t_s$. The basic version, where $jump = False$, returns $\mathcal{T}$ immediately (lines~\ref{alg2:return1}--\ref{alg2:return2}). While this version minimizes makespan, it can be inefficient if conflicts arise repeatedly with each shift in the while loop (lines~\ref{alg2:while_s}--\ref{alg2:while_e}), causing exponential growth in the search tree as \textsc{ConflictCheck} repeatedly detects conflicts (line~\ref{alg1:conflict}).


A simple way to reduce inefficiency is to wait longer for other robots to move. The challenge is determining the optimal wait time. If $jump = True$ (line~\ref{alg2:jump}), \textsc{SearchJumpStep} uses binary search to find $t_c$, calculating the pause duration as $t_c - t_s$. The trajectory is then shifted, and the empty slots are filled with $q$.


The binary search first checks the last time step $\tau_i$ of $\mathcal{T}.R$ for conflicts. If detected, it assumes conflicts persist until $t_c = \tau_i$ and waits until $t_c$. If no conflict is found, the next midpoint $t_s + 1 + \lfloor(\tau_i - t_s)/2\rfloor$ is checked. If no conflict appears, the search moves downward; if found, it moves upward to pinpoint when the conflict disappears.

\begin{figure}[t]
    \centering
    \captionsetup{skip=0pt}
    \includegraphics[width=0.4\textwidth]{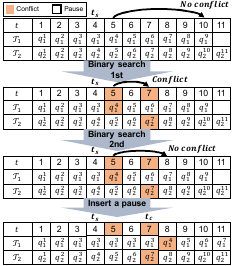}
    \caption{An example of \textsc{SearchJumpStep}: After resolving a conflict at $t_s = 4$, a binary search checks $q_1^4$ at $t_s + 1$ with $q_2^{10}$ at $t = 10$, finding no conflict. Next, it checks $q_2^7$ at $t = 7$, finding a conflict. The final check at $t = 8$ finds no conflict, setting $t_c$. A pause is then inserted at $q_1^3$ from $t_s = 4$ to $t_c = 7$.}
    \label{fig:binary}\vspace{-15pt}
\end{figure}

An example of the search is shown in Fig.~\ref{fig:binary}. After the while loop (lines~\ref{alg2:while_s}--\ref{alg2:while_e}), all conflicts are resolved up to $t_s$. If $jump = False$, Alg.~\ref{alg:stopngo} resumes and detects the next conflict after $t_s$ (line~\ref{alg1:conflict}). However, most conflicts occur in a row as manipulators are with a series of links that could block the other robot in conflict from proceeding. 
If $q^\star = q_1^4$ is in conflict at $t=5$, the trajectory is shifted by $1$, and $q_1^3$ is inserted at $t=5$ to add a pause. A conflict check is done at $t_s + 1$ and $\tau_1 = 10$. No conflict is found, so the next check is at the midpoint $t=7$. Finding a conflict there, the binary search moves upward to $t=8$, where no conflict is found, and the pause is inserted up to $t_c = 7$ ($R_1$ remains at $q_1^3$).


While this version significantly speeds up the basic version, it may miss non-conflicting steps between $t_s + 1$ and $t_c$, leading to unnecessary delays. Users can choose between the basic and binary search versions to balance solution optimality and planning time.


\subsection{Analysis of Stop-N-Go algorithm}

We show the basic version is optimal with an admissible heuristic and that both versions are probabilistically complete.

\thm \textbf{1} Given an initial collision-free trajectories $\mathcal{T}_{\text{init}}$, the basic version of Alg.~\ref{alg:stopngo} finds an optimal solution. 

\pf. The heuristic $h(\mathcal{N})$ in the basic version is the makespan of the conflict-free solution at node $\mathcal{N}$, based on resolving conflicts between a pair of robots. It does not account for potential conflicts involving other robots.


Since the heuristic function only considers the current makespan without taking into account future conflicts, it is a lower bound on the true makespan. If there are additional conflicts with other robots, the true makespan will likely be extended. Therefore, the heuristic function never overestimates the actual makespan so admissible. Since A$^*$ search with an admissible heuristic is guaranteed to find the optimal solution, the basic version of Alg.~\ref{alg:stopngo} is optimal.\qed

On the other hand, the heuristic function in the binary search version could overestimate the actual makespan as it could miss some non-conflicting steps as the binary search does not fully examine the whole trajectory. Note that the basic version guarantees an optimal solution for the given initial trajectories, but not necessarily for the entire motion planning instance, as different initial trajectories could yield a smaller makespan.


\thm \textbf{2} Alg.~\ref{alg:stopngo} is probabilistically complete as the probability that it finds a solution (if one exists) approaches $1$ as the number of samples approaches infinity. This holds for both the basic and the binary search version of \textsc{ConflictResolution}.

\pf. Once $\mathcal{T}_{\text{init}}$ is found, Alg.~\ref{alg:stopngo} only adjusts timing, not spatial configuration. Since the initial trajectories are collision-free with static obstacles, the remaining conflicts are from dynamic interactions as the robots move.


Alg.~\ref{alg:stopngo} systematically resolves such conflicts by inserting pauses so ensures that each robot waits until it is safe to proceed. A conflict-free solution must be found because the robots will eventually return to their initial configuration (the assumption in Sec.~\ref{sec:prob}) which is identical to the environment where $\mathcal{T}_{\text{init}}$ is found. In $\mathcal{T}_{\text{init}}$, the trajectory of each robot is found by assuming the other robots in their initial configurations. Thus, executing each of trajectories in $\mathcal{T}_{\text{init}}$ in turn will produce a conflict-free solution. In other words, $\mathcal{T}_i+1$ is executed after $\mathcal{T}_i$ is executed for all $i = 1, \cdots, k-1$ (i.e., equivalent to the sequential execution of all trajectories). This worst-case solution is guaranteed to be found so the search is complete. The binary search version might miss some non-conflicting steps so the robots could wait longer than the optimal solution. Nevertheless, the variation does not affect the completeness because Alg.~\ref{alg:stopngo} is guaranteed to find the worst-case conflict-free solution which is the sequential execution.

The initial trajectories in $\mathcal{T}_{\text{init}}$ are found using a sampling-based motion planner, which is probabilistically complete. Since the motion planning is probabilistically complete and the A$^*$ search is complete, the probability of finding a set of conflict-free trajectories approaches $1$. \qed


\section{Experiments}

\begin{figure*}[t]
    \centering
    \captionsetup{skip=0pt}
    \begin{subfigure}{0.95\textwidth}
        \centering
        \captionsetup{skip=0pt}
        \begin{subfigure}{0.329\textwidth}
            \includegraphics[width=\textwidth]{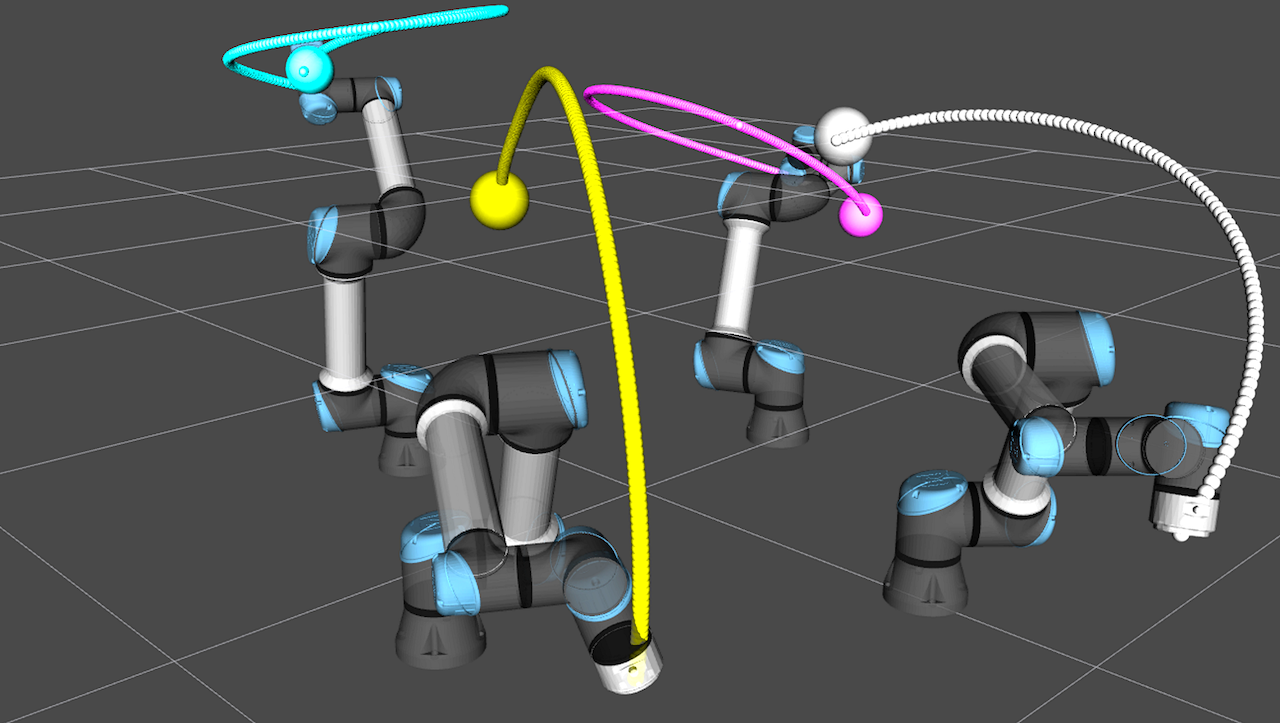}
        \end{subfigure}\,%
        \begin{subfigure}{0.276\textwidth}
            \includegraphics[width=\textwidth]{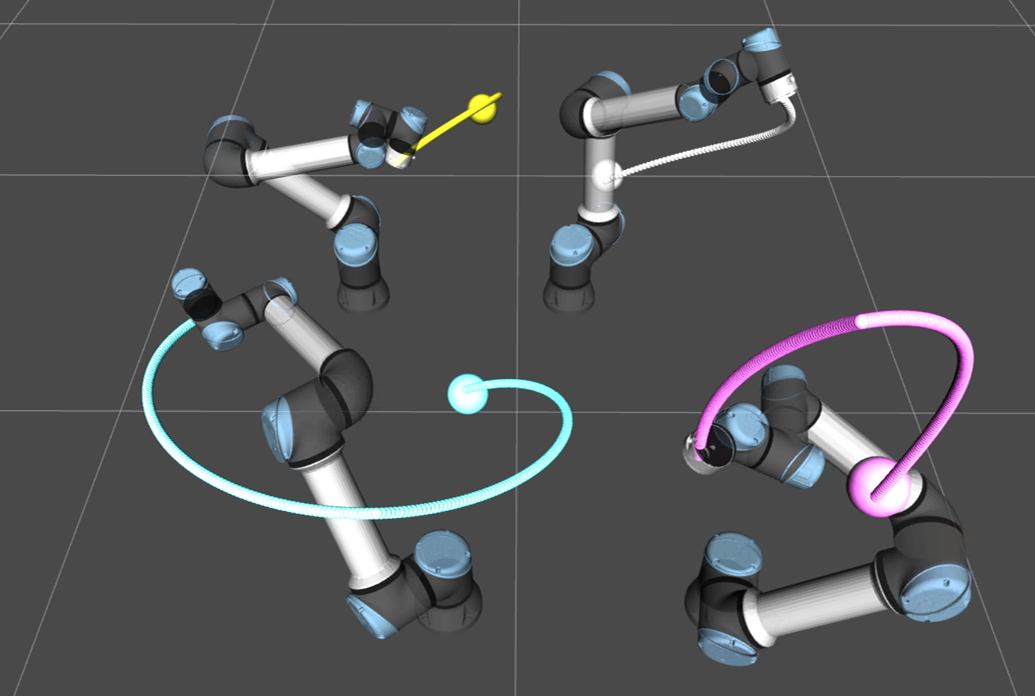}
        \end{subfigure}\,%
        \begin{subfigure}{0.362\textwidth}
            \includegraphics[width=\textwidth]{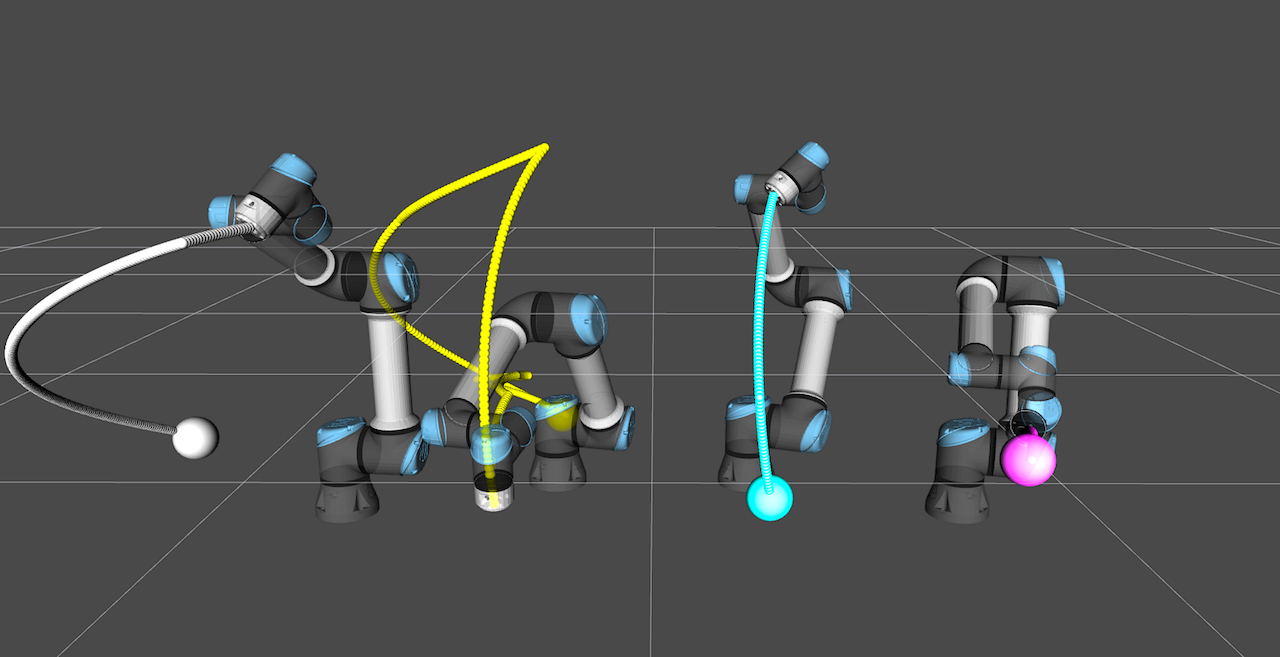}
        \end{subfigure}
        \caption{Examples of \textsf{Unbounded} where goal positions are sampled within the entire workspace of each robot}
        \label{fig:random_goal}
    \end{subfigure}\hfill%
    \begin{subfigure}{0.95\textwidth}
        \centering
        \captionsetup{skip=0pt}
        \begin{subfigure}{0.329\textwidth}
            \includegraphics[width=\textwidth]{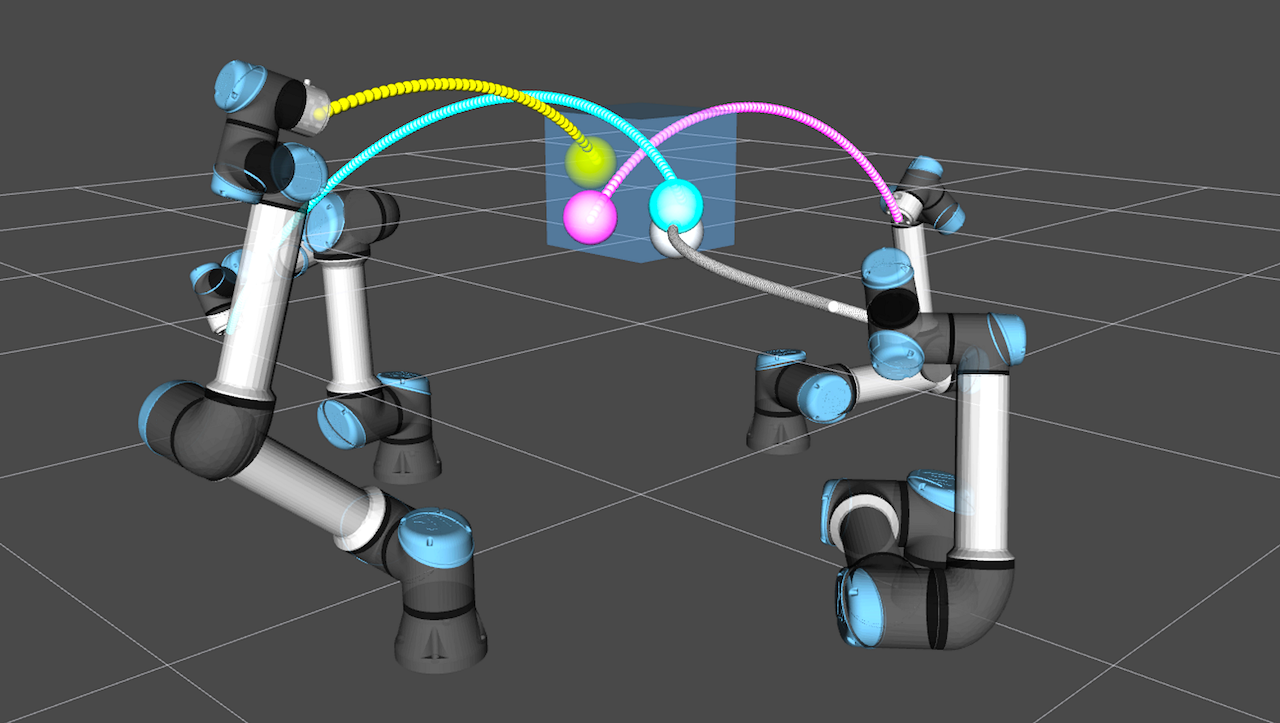}
        \end{subfigure}\,%
        \begin{subfigure}{0.276\textwidth}
            \includegraphics[width=\textwidth]{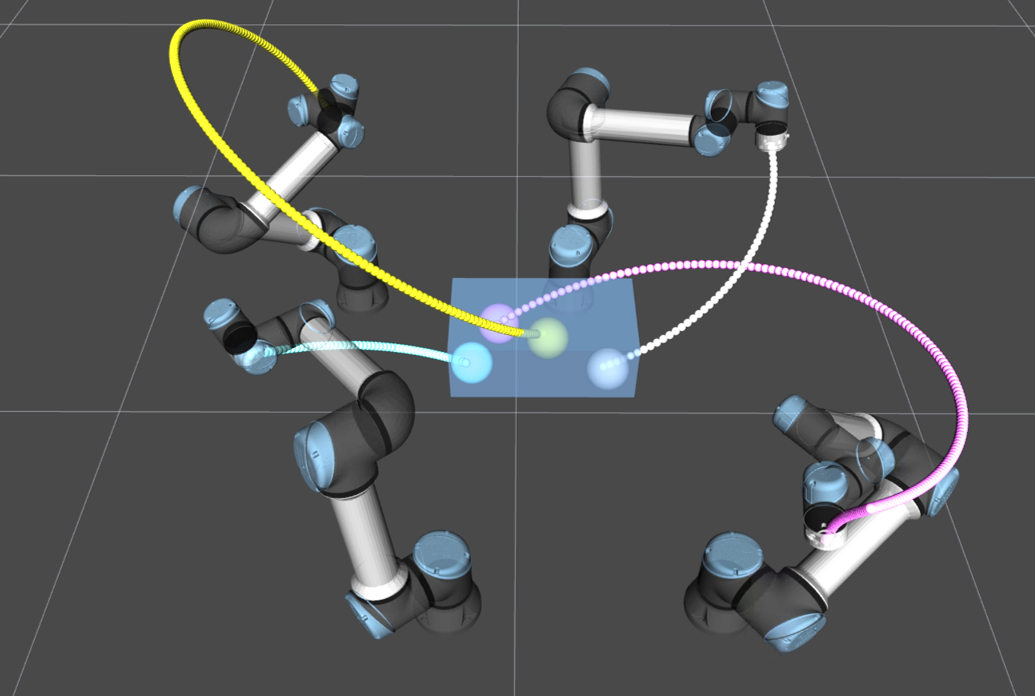}
        \end{subfigure}\,%
        \begin{subfigure}{0.362\textwidth}
            \includegraphics[width=\textwidth]{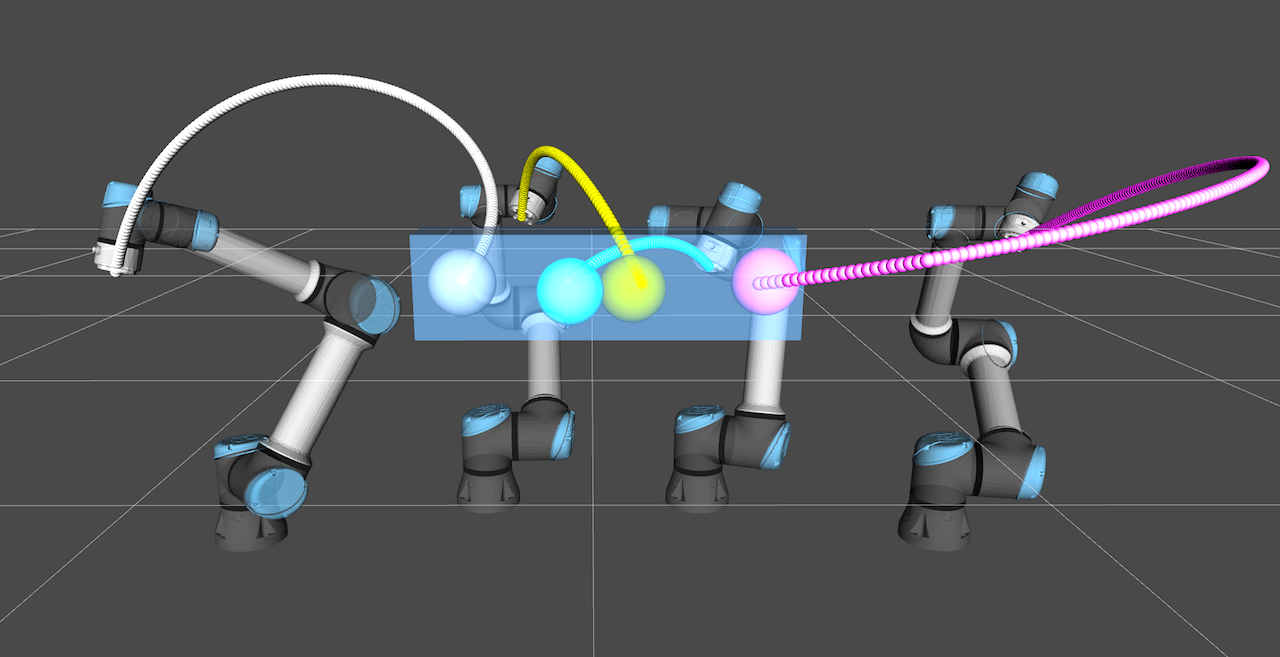}
        \end{subfigure}
        \caption{Examples of \textsf{Bounded} where goal positions are sampled from the shared workspace of the robots}
        \label{fig:bounded_goal}
    \end{subfigure}
    \caption{From the left, different arrangements of robots \textit{Square}, \textit{Zigzag}, and \textit{Trapezoid} are shown where (a) and (b) have different goal distributions.}\vspace{-10pt}
    \label{fig:env}
\end{figure*}

The advantage of our method is the ability to solve the challenging instances which require a temporal coordination between the trajectories, which often arise when multiple manipulators share a workspace with closely located goal positions. Fig.~\ref{fig:env} illustrates such the environments, also used in our experiments. The robots are arranged in three patterns: \textsf{Square}, \textsf{Zigzag}, and \textsf{Trapezoid}, with four robots in each setup. While our method can handle more manipulators, four are chosen as adding more robots in a tight task space is often impractical. We test two goal distributions: \textsf{Unbounded}, with random goal positions within the workspace of each robot (Fig.~\ref{fig:random_goal}), and \textsf{Bounded}, with goals clustered in a small cuboid within the shared workspace (Fig.~\ref{fig:bounded_goal}). \textsf{Bounded} presents more difficult scenarios due to the proximity of the goal positions.




The performance metrics are success rate, planning time, makespan, and total time. Makespan measures the time from the start until all robots complete their tasks and return. The total time sums the planning time and makespan so measures the entire time for a team of robots to solve a planning instance and execute to perform their tasks.
For six combinations of environments and goal distributions, we generate 15 random instances each, with random initial and goals configurations. Same sets of instances are used for comparison. The success rate is based on whether a solution is found within $40$ seconds ($10$ seconds for initial planning, $30$ seconds for A$^*$ search). We run other three methods for comparison: \textit{Sequential}, which executes $\mathcal{T}_\text{init}$ one by one, \textit{Coupled}, which plans in composite C-space, and a reinforcement learning method~\cite{ha2020multiarm} (\textit{RL}). The compared methods also have $40$ seconds for generating motions.

The test is done with four Universal Robots UR5 which is with 6 DOFs. Our method, \textit{Sequential}, and \textit{Coupled} use RRT-connect~\cite{RRTconnect} implemented in Open Motion Planning Library~\cite{OMPL} within MoveIt!~\cite{MoveIt}. Our method uses FCL~\cite{FCL} integrated with Assimp (Open Asset Import Library) for collision checking. The parameter for the trajectory interpolation is $t_\text{intv} = 0.3$. A small $t_{\text{intv}}$ leads to more collision checks so longer computation time while a large value could miss some collisions. Depending on the computational resources or the time allowed for planning, users could tune the value. \textit{RL} is executed using PyBullet simulator. All experiments are done in a system with an AMD 5800X 3.8GHz CPU, 32GB RAM, and C++11.

\begin{figure}[t]
    \centering
    \captionsetup{skip=0pt}
    \includegraphics[width=0.95\columnwidth]{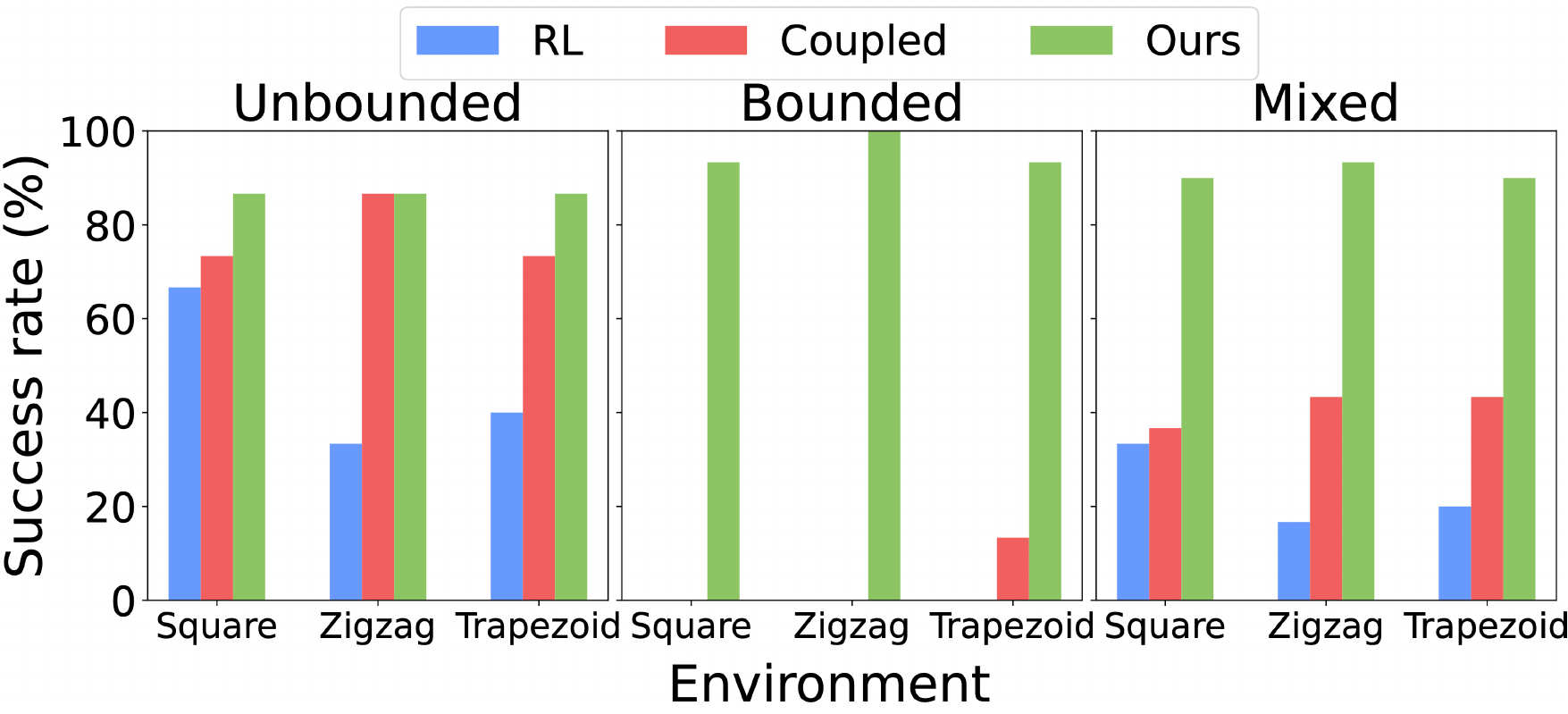}
    \caption{The success rates of different methods across three test environments with different distributions of the goal positions}    
    \label{fig:success_rate}\vspace{-10pt}
\end{figure}

The success rates are shown in Fig.~\ref{fig:success_rate} and Table~\ref{tab:success_rate}. In \textsf{Unbounded}, our method achieves higher success rates compared to \textit{Coupled} and \textit{RL}. We omit the result of \textit{Sequential} which succeeds always according to the nature of the method. In these instances, the goal positions are sparsely located so the initial trajectories often do not have conflicts. Thus, \textit{Coupled} also achieves the success rates above $50\%$. The success rate of \textit{RL} varies according to the arrangement of the robots. In \textit{Square}, the workspaces of the robots overlap only partially so there are only few or no conflict to be resolved. In such environments, \textsf{RL} performs well. However, in the environments where the overlap is nonnegligible such as \textit{Zigzag} and \textit{Trapezoid}, \textsf{RL} is prone to fail to find a solution. 

In \textsf{Bounded}, our method demonstrates exclusively higher success rates (above $90\%$) compared to others that rarely succeed. The main reason is that the test instances have goal positions within a small bounding box which incur the situations where temporal adjustments are critical. For a mix of the test instances of \textsf{Unbounded} and \textsf{Bounded}, our method achieves significantly higher success rates compared to others, supporting that our method is able to plan trajectories for multiple high-DOF manipulators effectively and efficiently. Since our method is proven to be probabilistically complete, few failing instances represent that there are no initial trajectories found owing to an insufficient number of samples or some random goals are unreachable.

\begin{table}[t]
\captionsetup{skip=0pt}
\caption{The success rates of different methods across three test environments with different distributions of the goal positions. Bold numbers represent the best result.}
\label{tab:success_rate}
\centering
\resizebox{0.95\columnwidth}{!}{%
\begin{tabular}{|c|c||c|c|c|}
\hline
\multirow{2}{*}{Environment} & \multirow{2}{*}{Method} & \multicolumn{3}{c|}{Success rate (\%)} \\ \cline{3-5}
&           & Unbounded         & Bounded           & Total          \\ \hline

\multirow{3}{*}{Square} 
 & RL       & 66.67             & 0                 & 33.33         \\ \cline{2-5}
 & Coupled  & 73.33             & 0                 & 36.67         \\ \cline{2-5}
 & Ours     & \textbf{86.67}    & \textbf{93.33}    & \textbf{90}   \\ \hline
 
\multirow{3}{*}{Zigzag} 
 & RL       & 33.33             & 0                 & 16.67         \\ \cline{2-5}
 & Coupled  & \textbf{86.67}    & 0                 & 43.33         \\ \cline{2-5}
 & Ours     & \textbf{86.67}    & \textbf{100}      & \textbf{93.33} \\ \hline
 
\multirow{3}{*}{Trapezoid} 
 & RL       & 40                & 0                 & 20            \\ \cline{2-5}
 & Coupled  & 73.33             & 13.33             & 43.33         \\ \cline{2-5}
 & Ours     & \textbf{86.67}    & \textbf{93.33}    & \textbf{90}   \\ \hline
 
\end{tabular}}\vspace{-10pt}
\end{table}

\begin{figure}[t]
    \centering
    \captionsetup{skip=0pt}
    \includegraphics[width=0.9\columnwidth]{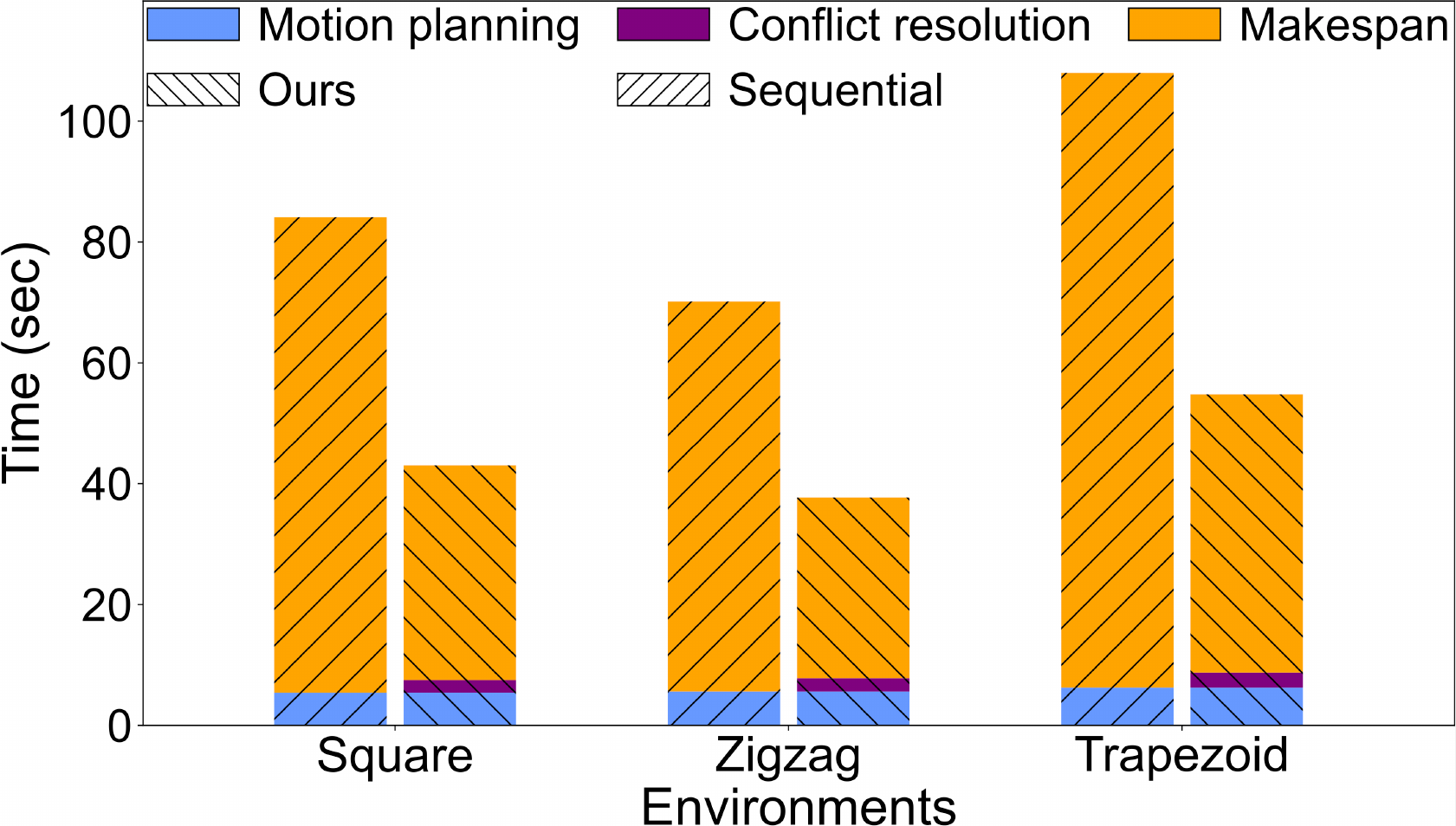}
    \caption{The planning and execution time of \textit{Sequential} and our method. The time is further broken down into motion planning time, search time for conflict resolution, and makespan. This plot shows the averages over the results from \textsf{Bounded}  and \textsf{Unbounded}. Statistics are shown in Table~\ref{tab:makespan}. }
    \label{fig:makespan}\vspace{-10pt}
\end{figure}

The next comparison focuses on the quality of the solutions found (i.e., makespan). The result is shown in Fig.~\ref{fig:makespan} and Table~\ref{tab:makespan}. Since \textit{Coupled} and \textit{RL} seldom succeed, their statistics may be unreliable. For instance, \textit{Coupled} in \textsf{Bounded} has only $2$ successful instances from which to compute the average. We also note that the compared methods tend to succeed in easy instances but fail in more challenging ones, resulting in a shorter makespan. In \textsf{Unbounded}, \textit{RL} exhibits the shortest makespan. Since \textit{RL} does not require the time for planning, its makespan includes the execution time only. However, it is not guaranteed to find a solution as shown by its success rate below $50\%$ and smaller than other compared methods. 
In \textsf{Bounded}, it fails to find a solution in any of the instances. Also, it requires (i) a long training time, consisting of $700,000$ episodes which takes roughly $14$ days on a system with NVIDIA GTX 1080, and (ii) the generation of $1,000,000$ expert waypoints taking significant time. These hurdles for training along with the incompleteness of the method are major drawbacks.

On the other hand, \textit{Coupled} shows shorter makespan with easy instances in \textsf{Unbounded} although its success rates are not competitive. With difficult instances in \textsf{Bounded}, it cannot succeed in finding a solution in all instances. While \textit{Sequential} is guaranteed to find a solution if exists, its makespan is several times longer than ours. 

Overall, the experimental results show that our method is able to solve multi-manipulator motion planning instances that cannot be solved by \textit{Coupled} and \textit{RL}. Compared to the complete method \textit{Sequential} given initial trajectories, our method shows significantly shorter makespans (Fig.~\ref{fig:makespan}).

\begin{table}[t]
\centering
\captionsetup{skip=0pt}
\caption{Planning and execution time for (a) \textsf{Unbounded} and (b) \textsf{Bounded}. Standard deviations are indicated in parentheses.}
\label{tab:makespan}
\begin{subtable}[t]{0.49\textwidth}
    \centering
    \captionsetup{skip=0pt}
        \caption{Unbounded}
    \resizebox{0.99\columnwidth}{!}{%
    \begin{tabular}{|c|c||c|c|c|c|}
    \hline
    \multirow{3}{*}{Environment} & \multirow{3}{*}{Method} & \multicolumn{4}{c|}{Time (sec)} \\ \cline{3-6}
    &               & \multicolumn{1}{c|}{\makecell{Motion\\ planning}}         & \multicolumn{1}{c|}{\makecell{Conflict\\ resolution}}           & Makespan & Total          \\ \hline

    \multirow{6}{*}{Square} 
     & Sequential   & \makecell{\textbf{4.67}\\ \textbf{(3.07)}}    & -                         & \makecell{70.8\\ (25.37)}                     & \makecell{75.46\\ (25.65)}    \\ \cline{2-6}
     & Coupled      & \makecell{5.96\\ (4.61)}                      & -                         & \makecell{15.07\\ (5.91)}                     & \makecell{21.03\\ (5.57)}     \\ \cline{2-6}
     & RL           & -                                             & -                         & \makecell{\textbf{7.25}\\ \textbf{(1.22)}}    & \makecell{\textbf{7.25}       \\ \textbf{(1.22)}}  \\ \cline{2-6}
     & Ours         & \makecell{\textbf{4.67}\\ \textbf{(3.07)}}    & \makecell{0.43\\ (0.18)}  & \makecell{34.03\\ (14.95)}                    & \makecell{39.12\\ (15.1)}     \\ \hline

    \multirow{6}{*}{Zigzag} 
     & Sequential   & \makecell{5.58\\ (3.63)}                     & -                          & \makecell{58.64\\ (21.52)}                    & \makecell{64.22\\ (22.04)}    \\ \cline{2-6}
     & Coupled      & \makecell{\textbf{4.04}\\ \textbf{(3.997)}}  & -                          & \makecell{14.36\\ (7.68)}                     & \makecell{18.41\\ (7.55)}     \\ \cline{2-6}
     & RL           & -                                            & -                          & \makecell{\textbf{7.29}\\ \textbf{(1.41)}}    & \makecell{\textbf{7.29}       \\ \textbf{(1.41)}}  \\ \cline{2-6}
     & Ours         & \makecell{5.58\\ (3.63)}                     & \makecell{0.63\\ (0.8)}    & \makecell{26.32\\ (10.18)}                    & \makecell{32.55\\ (11.26)}    \\ \hline

    \multirow{6}{*}{Trapezoid} 
     & Sequential   & \makecell{\textbf{4.9}\\ \textbf{(2.55)}}    & -                          & \makecell{92.09\\ (30.5)}                     & \makecell{96.99\\ (31.5)}     \\ \cline{2-6}
     & Coupled      & \makecell{7.87\\ (6.93)}                     & -                          & \makecell{23.79\\ (8.29)}                     & \makecell{31.66\\ (10.27)}    \\ \cline{2-6}
     & RL           & -                                            & -                          & \makecell{\textbf{7.89}\\ \textbf{(0.94)}}    & \makecell{\textbf{7.89}       \\ \textbf{(0.94)}}  \\ \cline{2-6}
     & Ours         & \makecell{\textbf{4.9}\\ \textbf{(2.55)}}    & \makecell{1.88\\ (2.59)}   & \makecell{43.38\\ (15.28)}                    & \makecell{48.21\\ (17.1)}     \\ \hline

    \end{tabular}}
    \label{tab:unbounded}
\end{subtable}\\
\vspace{3pt}
\begin{subtable}[t]{0.49\textwidth}
    \centering
    \captionsetup{skip=0pt}
        \caption{Bounded}
    \resizebox{0.99\columnwidth}{!}{%
    \begin{tabular}{|c|c||c|c|c|c|}
    \hline
    \multirow{3}{*}{Environment} & \multirow{3}{*}{Method} & \multicolumn{4}{c|}{Time (sec)} \\ \cline{3-6}
    &               & \multicolumn{1}{c|}{\makecell{Motion\\ planning}}         & \multicolumn{1}{c|}{\makecell{Conflict\\ resolution}}           & Makespan & Total          \\ \hline

    \multirow{3}{*}{Square} 
    & Sequential   & \multirow{3}{*}{\makecell{6.12\\ (4.16)}}  & -                         & \makecell{86\\ (29.02)}                       & \makecell{92.12\\ (32.28)}     \\ \cline{2-2} \cline{4-6}
    & Ours         &                                            & \makecell{3.56\\ (5.14)}  & \makecell{\textbf{36.97}\\ \textbf{(9.76)}}   & \makecell{\textbf{46.72}\\ \textbf{(13.41)}}     \\ \hline

    \multirow{3}{*}{Zigzag} 
    & Sequential   & \multirow{3}{*}{\makecell{5.6\\ (3.15)}}   & -                         & \makecell{71.42\\ (18.84)}                    & \makecell{77.02\\ (18.98)}     \\ \cline{2-2} \cline{4-6}
    & Ours         &                                            & \makecell{4.03\\ (6.54)}  & \makecell{\textbf{34.03}\\ \textbf{(10.05)}}  & \makecell{\textbf{46.65}\\ \textbf{(14.76)}}
     \\ \hline

    \multirow{5}{*}{Trapezoid}
    & Sequential   & \makecell{\textbf{7.65}\\ \textbf{(4.13)}} & -                         & \makecell{113.4\\ (36.34)}                    & \makecell{121.05\\ (32.72)}     \\ \cline{2-6}
    & Coupled      & \makecell{10.11\\ (8.73)}                  & -                         & \makecell{\textbf{27.46}\\ \textbf{(2.32)}}   & \makecell{\textbf{37.58}\\ \textbf{(6.4)}}   \\ \cline{2-6}
    & Ours         & \makecell{\textbf{7.65}\\ \textbf{(4.13)}} & \makecell{3.01\\ (4.07)}  & \makecell{50.31\\ (14.69)}                    & \makecell{56.74\\ (15.02)}     \\ \hline
    \end{tabular}}
    \label{tab:bounded}
\end{subtable}
\end{table}

\section{Conclusion}

We propose Stop-N-Go, a motion planning method for multiple manipulators in densely packed environments. Conflicts between initial trajectories are resolved by inserting pauses, with solutions found via A$^*$ search. Our method is probabilistically complete, optimal, and achieves higher success rates than methods that overlook the temporal dimension. Experiments demonstrate its effectiveness, particularly when robots have closely located goals. We are currently developing a physical system with two manipulators to test this approach, with results to be presented in future work.


\clearpage
\bibliographystyle{IEEEtran}
\bibliography{ref}

\begin{thebibliography}{10}
\providecommand{\url}[1]{#1}
\csname url@rmstyle\endcsname
\providecommand{\newblock}{\relax}
\providecommand{\bibinfo}[2]{#2}
\providecommand\BIBentrySTDinterwordspacing{\spaceskip=0pt\relax}
\providecommand\BIBentryALTinterwordstretchfactor{4}
\providecommand\BIBentryALTinterwordspacing{\spaceskip=\fontdimen2\font plus
\BIBentryALTinterwordstretchfactor\fontdimen3\font minus \fontdimen4\font\relax}
\providecommand\BIBforeignlanguage[2]{{%
\expandafter\ifx\csname l@#1\endcsname\relax
\typeout{** WARNING: IEEEtran.bst: No hyphenation pattern has been}%
\typeout{** loaded for the language `#1'. Using the pattern for}%
\typeout{** the default language instead.}%
\else
\language=\csname l@#1\endcsname
\fi
#2}}

\bibitem{RRTconnect}
J.~J. Kuffner and S.~M. LaValle, ``Rrt-connect: An efficient approach to single-query path planning,'' in \emph{Proceedings 2000 ICRA. Millennium Conference. IEEE International Conference on Robotics and Automation. Symposia Proceedings (Cat. No. 00CH37065)}, vol.~2.\hskip 1em plus 0.5em minus 0.4em\relax IEEE, 2000, pp. 995--1001.

\bibitem{dRRT}
R.~Shome, K.~Solovey, A.~Dobson, D.~Halperin, and K.~E. Bekris, ``drrt*: Scalable and informed asymptotically-optimal multi-robot motion planning,'' \emph{Autonomous Robots}, vol.~44, no.~3, pp. 443--467, 2020.

\bibitem{PP}
M.~Erdmann and T.~Lozano-Perez, ``On multiple moving objects,'' \emph{Algorithmica}, vol.~2, pp. 477--521, 1987.

\bibitem{stilman2005navigation}
M.~Stilman and J.~J. Kuffner, ``Navigation among movable obstacles: Real-time reasoning in complex environments,'' \emph{International Journal of Humanoid Robotics}, vol.~2, no.~04, pp. 479--503, 2005.

\bibitem{PRM}
L.~E. Kavraki, P.~Svestka, J.-C. Latombe, and M.~H. Overmars, ``Probabilistic roadmaps for path planning in high-dimensional configuration spaces,'' \emph{IEEE Transactions on Robotics and Automation}, vol.~12, no.~4, pp. 566--580, 1996.

\bibitem{CBS}
G.~Sharon, R.~Stern, A.~Felner, and N.~R. Sturtevant, ``Conflict-based search for optimal multi-agent pathfinding,'' \emph{Artificial Intelligence}, vol. 219, pp. 40--66, 2015.

\bibitem{ha2020multiarm}
H.~Ha, J.~Xu, and S.~Song, ``Learning a decentralized multi-arm motion planner,'' in \emph{Conference on Robotic Learning (CoRL)}, 2020.

\bibitem{soft_actor_critic}
T.~Haarnoja, A.~Zhou, P.~Abbeel, and S.~Levine, ``Soft actor-critic: Off-policy maximum entropy deep reinforcement learning with a stochastic actor,'' in \emph{International conference on machine learning}.\hskip 1em plus 0.5em minus 0.4em\relax PMLR, 2018, pp. 1861--1870.

\bibitem{BiRRT}
A.~H. Qureshi and Y.~Ayaz, ``Intelligent bidirectional rapidly-exploring random trees for optimal motion planning in complex cluttered environments,'' \emph{Robotics and Autonomous Systems}, vol.~68, pp. 1--11, 2015.

\bibitem{FCL}
J.~Pan, S.~Chitta, and D.~Manocha, ``Fcl: A general purpose library for collision and proximity queries,'' in \emph{2012 IEEE International Conference on Robotics and Automation}.\hskip 1em plus 0.5em minus 0.4em\relax IEEE, 2012, pp. 3859--3866.

\bibitem{OMPL}
I.~A. Sucan, M.~Moll, and L.~E. Kavraki, ``The open motion planning library,'' \emph{IEEE Robotics \& Automation Magazine}, vol.~19, no.~4, pp. 72--82, 2012.

\bibitem{MoveIt}
A.~S. Ioan and S.~Chitta, ``Moveit,'' \emph{Available online: moveit. ros. org.(accessed on 2 September 2021)}, 2019.

\end{thebibliography}

\end{document}